\documentclass[letterpaper, 10 pt, conference]{ieeeconf}

\IEEEoverridecommandlockouts   

\usepackage{dsfont}
\usepackage[T1]{fontenc}
\usepackage{graphicx, stfloats} 
\usepackage{svg}
\usepackage{geometry}
\usepackage{standalone}
\usepackage{hyperref}
\usepackage{pgfplots}
\usepackage{booktabs}
\usepackage{subcaption}
\usepackage{amssymb, amsmath}
\usepackage{algorithm}
\usepackage{algpseudocode}
\usepackage{makecell}
\usepackage{bbm}
\usepackage{diagbox, multirow} 
\usepackage{color,soul} 
\usepackage{siunitx}

\newcommand{\revision}[1]{\textcolor{black}{ #1}}

\hypersetup{
    colorlinks=true, 
    allcolors=black, 
    pdfborder={0 0 0} 
}

\definecolor{vandeusen}{RGB}{73,92,111}
\definecolor{cordovan}{RGB}{152,68,71}
\definecolor{alizarin}{rgb}{0.82, 0.1, 0.26}
\definecolor{azure}{rgb}{0.0, 0.5, 1.0}

\usepackage[style=ieee, url=false,doi=false,eprint=false,natbib=false,mincitenames=1,maxcitenames=1,dashed=false,isbn=false, maxbibnames=99]{biblatex}

\title{\LARGE \bf Importance Sampling-Guided Meta-Training for Intelligent Agents in Highly Interactive Environments}

\author{ Mansur M. Arief$^{1\star}$, Mike Timmerman$^{1\star}$, Jiachen Li$^2$, David Isele$^3$, Mykel J. Kochenderfer$^1$%
\thanks{$^{1}$Department of Aeronautics and Astronautics, Stanford University  (e-mail: \{ariefm, mtimmerman, mykel\}\!@stanford.edu).}%
\thanks{$^{2}$Department of Electrical and Computer Engineering, University of California, Riverside (e-mail: jiachen.li@ucr.edu)}%
\thanks{$^{3}$Honda Research Institute (e-mail: disele@honda-ri.com)}%
\thanks{$^{\star}$Indicates equal contribution}%
}

\begin{document}

\maketitle

\begin{abstract}
Training intelligent agents to navigate highly interactive environments presents significant challenges. While guided meta reinforcement learning (RL) approach that first trains a guiding policy to train the ego agent has proven effective in improving generalizability across scenarios with various levels of interaction, the state-of-the-art method tends to be overly sensitive to extreme cases, impairing the agents' performance in the more common scenarios. 
This study introduces a novel training framework that integrates guided meta RL with importance sampling (IS) to optimize training distributions iteratively for navigating highly interactive driving scenarios\revision{, such as T-intersections or roundabouts.} 
Unlike traditional methods that may underrepresent critical interactions or overemphasize extreme cases during training, our approach strategically adjusts the training distribution towards more challenging driving behaviors using IS proposal distributions and applies the importance ratio to de-bias the result. 
By estimating a naturalistic distribution from real-world datasets and employing a mixture model for iterative training refinements, the framework ensures a balanced focus across common and extreme driving scenarios. 
Experiments conducted with both synthetic and naturalistic datasets demonstrate both accelerated training and performance improvements under highly interactive driving tasks.
\end{abstract}

\section{Introduction}

Autonomous agents often navigate complex, highly interactive environments, such as congested and unsignalized intersections \cite{isele2018navigating,wei2021autonomous,ma2021reinforcement,li2024multi,anon2024multi,li2024interactive}. The direct training of these agents using a naturalistic distribution of driving scenarios is notably inefficient due to the imbalanced frequency of scenarios---common scenarios are overrepresented while critical, interactive scenarios are rare yet essential for training \cite{arief2018accelerated, ding2023survey}.

It is an open question of how to best train agents when the distribution associated with their target operation is unknown \cite{schaul2015prioritized,isele2018selective}. Recent developments have highlighted the effectiveness of guided meta reinforcement learning (RL) in training agents to handle a variety of interactive driving situations \cite{lee2023robust}. Here, the core strategy involves training the ego agent to interact with a range of social agents, each governed by uniquely parameterized reward functions. These functions are designed to emulate a broad spectrum of cooperative or adversarial behaviors, thus aiming to mimic human-like driving patterns to increase the robustness of the ego agent's policy against naturalistic driving conditions. Nonetheless, this method tends to overemphasize extreme scenarios that might not be as frequent in real-world driving but are sampled disproportionately during training, leading to performance degradation for more common driving conditions \cite{lee2023robust,li2024adaptive}.

To address this issue, strategies such as oversampling and undersampling have been discussed in the literature \cite{zhou2020oversampling, hong2023beyond}. Implementing these methods in an online training environment remains challenging. Techniques from variance reduction, such as using importance sampling (IS) for efficient evaluation sampler have been proposed. The main idea of IS in this scheme is to skew the sampling towards extreme scenarios and then adjust the outcomes with importance ratios for an unbiased result. While IS-based approaches have shown appealing results, these approaches are typically applied in evaluation phases \cite{zhao2016accelerated,o2018scalable,arief2022certifiable}. Whether these approaches could be adapted to enhance training remains largely underexplored.

\begin{figure} 
    \centering
    \includegraphics[width=\linewidth]{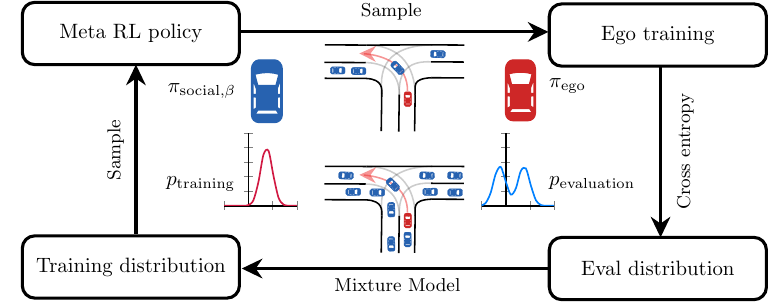}
    \caption{The IS-guided meta RL training framework operates as follows. First, we train a meta RL policy that captures the diverse behaviors of social agents, characterized by varying levels of aggressiveness ($\beta$). Next, the ego agent is trained using samples of these behaviors from a distribution $p_{\text{training}}$. Its performance is then evaluated using a cross-entropy IS proposal $p_{\text{evaluation}}$, which emphasizes unresolved failure modes. Finally, we create a mixture model from both $p_{\text{training}}$ and $p_{\text{evaluation}}$ for use in the next iteration of training.\label{fig:framework}}
    \vspace{-0.3cm}
\end{figure}

This study introduces a framework that employs IS both during training and evaluation to mitigate the challenge of overemphasis on extreme scenarios. The proposed training framework integrates a guided meta RL training approach with IS, optimizing the training distribution to efficiently sample critical interactive scenarios without disproportionately emphasizing them. 
The IS-optimized training approach strategically biases sampling towards more aggressive driving situations using an IS proposal derived through the cross-entropy method~\cite{kroese2013cross}, then computes the importance ratio based on this proposal and the underlying naturalistic distribution to provide an unbiased reward estimate. We validate our approach on \revision{T-intersection~\cite{inDdataset} and roundabout~\cite{rounDdataset} scenarios}, demonstrating that it not only enhances the training process but also maintains a balanced emphasis on common and critical cases.

In summary, our contributions are threefold:
\begin{itemize}
    \item We integrate the IS proposal, often used for evaluation, into an optimized training distribution.
    \item We introduce an IS-guided meta RL policy training framework that effectively balances the training for both common and critical driving scenarios, enhancing the training process by reflecting naturalistic driving dynamics more accurately. 
    \item We demonstrate the generalizability of the proposed approach in real-world driving datasets, which achieves significant improvement compared to baselines.
\end{itemize}

The remainder of the paper is structured as follows. Section~\ref{sec:related_work} provides a brief review of the related work. Section~\ref{sec:framework} details our methodological approach, followed by Section~\ref{sec:experiments} that describes our experimental settings. Section~\ref{sec:discussion} discusses our findings, concluding with Section~\ref{sec:conclusion} that summarizes our study.

\section{Related Work}\label{sec:related_work}


\subsection{Policy Learning for Autonomous Navigation}

A broad spectrum of planning and policy learning frameworks have been proposed, evolving from earlier models such as simple Markov Decision Processes (MDPs) \cite{guan2018markov} to more sophisticated structures like Partially Observable MDPs (POMDPs) \cite{qiao2018pomdp, sunberg2022improving} and deep RL methods \cite{zhu2021survey, li2024interactive}. These frameworks have contributed significantly to the development of autonomous driving policies by enabling the modeling of uncertainty and partial observability, which are crucial in dynamic driving environments. Training diverse policies under these frameworks is achieved through reward randomization techniques \cite{tang2021discovering, mysore2021multi}, resulting in policies capable of handling complex scenarios such as navigating through uncontrolled intersections. 

Despite these advancements, effectively addressing rare critical scenarios remains a challenge. Techniques such as conditional flows \cite{ding2023causalaf}, adaptive scenario-set sampling \cite{du2019finding, zhang2022finding}, and bridge sampling \cite{sinha2020neural} have been developed to enhance the sampling process. 
However, the complexity of these methods poses significant integration challenges when used during the training phase. There is also a persistent risk that policies may become excessively biased towards rare, extreme cases, potentially compromising the agent's overall performance in more common driving scenarios \cite{lee2023robust}.

\subsection{IS-based Evaluation Approaches}

Importance Sampling (IS) has been used to prioritize sampling in various applications \cite{ichter2018learning, luo2019importance} and has emerged as a crucial technique in evaluating the effectiveness of autonomous agent policies under rare and extreme scenarios. IS samples aggressive scenarios more frequently and applies likelihood ratios to ensure unbiased performance estimations \cite{kroese2013cross}. The aggressive scenario distribution used in IS, referred to as the IS proposal distribution, can be refined through methods like cross-entropy (CE) \cite{kroese2013cross}, dominating point analysis with mixture models  \cite{arief2022certifiable}, and failure-based normalizing flows \cite{gibson2023flow}.

While previous studies have shown the potential of addressing rare, extreme scenarios during planning \cite{kim2017heuristics}, a significant challenge remains on how to effectively bridge the gap between evaluation techniques and actual policy training. This disconnect often results in trained policies that do not fully align with real-world conditions. Effective integration of IS-based evaluation approaches with policy learning methods is crucial to ensure that trained policies are not only theoretically sound but also practical.

\subsection{Autonomous Agents in Highly Interactive Environments}
Designing autonomous agents capable of operating in highly interactive environments has been a major focus of recent research. Traditional approaches often rely on predefined rules and reactive behaviors, which are insufficient for handling the complexities and uncertainties present in real-world scenarios. Advanced frameworks such as interactive POMDPs ~\cite{doshi2005particle, luo2021interactive} and multi-agent reinforcement learning \cite{zhang2021multi, semnani2020multi} have been developed to address these challenges by considering the actions and behaviors of other agents in the environment.

Recent studies have also explored the use of social dynamics and negotiation strategies to enhance the interaction capabilities of autonomous agents. For instance, game theory can be used to model interactions among multiple agents allowing for the prediction and anticipation of other agents' actions, leading to more effective decision making processes \cite{bellemare2016unifying, yuan2021deep,li2023game}. Additionally, incorporating techniques such as imitation learning and behavior cloning has enabled agents to learn from human demonstrations, thereby improving their ability to interact seamlessly with human drivers and pedestrians \cite{hussein2017imitation, wang2021imitation}.

In this work, we design a framework that can robustly handle highly interactive and dynamic environments without sacrificing its nominal performance. Existing approaches often struggle with generalization, particularly when dealing with rare, critical interactions or highly unpredictable scenarios. Our research stands out by proposing a framework that integrates IS methods with meta RL to enhance the ability to navigate complex environments and effectively interact with other agents.

\section{Framework}\label{sec:framework}

The proposed framework uses an optimized IS proposal to enhance both the evaluation and subsequent training efficiency of autonomous driving agents. This dual application of IS is pivotal in generating critical scenarios that are not only essential for unbiased policy assessment but also improves the policy.

\subsection{Modeling and Objective}

We model the driving scenario as a partially observable stochastic game, where the interaction dynamics are described using the interactive driving model from \citeauthor{lee2023robust}~\cite{lee2023robust}. At any given time $t$, the scenario is defined by a state $s_t$. The objective for the ego policy $\pi^*_{\text{ego}}$ is to maximize its expected cumulative reward over time, formulated as:
\begin{align}
\pi^*_{\text{ego}} = \arg \max_{\pi \in \Pi_{\text{ego}}} \mathds{E} \left[ \sum_{t=0}^{\infty} \gamma^{t}R(s_t, \pi(s_t)) \right],
\end{align}
where $R$ represents the reward function for the ego vehicle, $\gamma$ is the discount factor, indicating the decreasing importance of future rewards, and the expectation is taken over state transitions. The ego policy $\pi$ maps the state space $\mathcal{S}$ to the action space $\mathcal{A}_{\text{ego}}$, with $\Pi_{\text{ego}}$ representing the feasible policy set for the ego agent.

\subsection{Social and Meta-Policy Training}

The social agents are modeled with a policy $\pi_{\text{social}}$, parameterized by $\beta$, which indicates the level of aggressiveness of the social agents. The policy for each social agent is optimized:
\begin{align}
    \max_{\pi \in \Pi_{\text{social}}} \mathds{E} \left[ \sum_{t=0}^\infty \gamma^t (\revision{r^{\text{1}}}(s_t, a_{\text{social}, t}) + \beta \revision{r^{\text{2}}}(s_t, a_{\text{social}, t})) \right],\label{eq:social_objective}
\end{align}
where $a_{\text{social}, t} = \pi_{\text{social}, \beta}(s_t)$, and \revision{and $r^{\text{1}}$ and $r^{\text{2}}$ are the primary and secondary reward components, respectively.}

To train a diverse set of social behaviors, we employ a meta-policy $\pi_{\text{social}, \beta}$ using a two-stage approach. In the first stage, baseline policies $\pi_{\text{social}, \beta}$ are trained for discrete preferences within a set $\overline{B} = \{\overline{\beta}^1, \ldots, \overline{\beta}^m\}$. \revision{Each baseline policy $\pi^*_{\text{social}, \overline{\beta}^i}$ targets a specific behavioral model parameterized by $\overline{\beta}_i$ and is a neural network that uses a set of optimized internal parameters that maximize its reward. These policies are called the guiding policies in the baseline method~\cite{lee2023robust}. The term `behavioral model' here refers to a driving policy that models a behavior induced by rewards defined by \autoref{eq:social_objective}, which, in more general sense, may consider various primary or secondary goals. In this work, however, we only consider the case for behavioral models $\pi_{\text{social}, \beta}$, $\pi_{\text{social}, \beta^\prime}$ for social vehicles that differ only due to $\beta \ne \beta^\prime$, where $\beta, \beta^\prime \in B \subseteq \mathds{R}$, but uses the same primary and secondary reward components.}

In the second stage, the meta-policy $\pi_{\text{social}, \beta}$ is then trained by sampling $\beta$ from a continuous distribution $U(\beta_{\text{min}}, \beta_{\text{max}})$, and is regularized to approximate the nearest \revision{guiding} policy using the regularization loss:
\begin{align}
    \mathcal{L}_{\text{reg}}(\pi_{\text{social}, \beta}) = \sum_{\overline{\beta} \in \overline{B}} \mathds{1}(|\overline{\beta} - \beta| \leq d) D_{KL}(\pi^*_{\text{social}, \overline{\beta}} \| \pi_{\text{social}, \beta}),\label{eq:reg_loss}
\end{align}
where $\mathds{1}$ is an indicator function, $D_{KL}$ is the Kullback-Leibler divergence, \revision{and $d$ is a small threshold. We also use $\beta_{\min} = \min_{\beta \in \overline{B}} \beta$ and $\beta_{\max} = \max_{\beta \in \overline{B}} \beta$. With this setting, the two-stage approach facilitates the synthesis of a diverse meta-policy capable of adapting to various social behaviors contained in $\overline{B}$, allowing the social policy to learn driving behaviors characterized by parameter $\beta$ and anchored at the guiding policies defined by $\overline{B}$.} Note that $\beta$ is only needed during training.

\subsection{Training the Ego Policy}

The ego policy $\pi_{\text{ego}}$ is trained against the backdrop of these diverse social policies. We consider several strategies for the training distribution of $\beta$, denoted by $p_{\text{training}}$, to prepare the ego policy for a spectrum of social behaviors.

\paragraph{Generalized Ego Policy (GEP)} Uses a uniform distribution $U(\beta_\text{min}, \beta_\text{max})$ for $p_\text{training}$ to prepare the ego policy for a wide range of social behaviors, potentially at the cost of overfitting to less common aggressive behaviors.

\paragraph{Generalized with IS (GIS)} Similar to GEP, but uses importance sampling weight to account for the rarity of the training scenarios.

\paragraph{Naturalistic Ego Policy (NEP)} Uses a distribution $p_\text{naturalistic}$ derived from real-world driving data to focus on common social behaviors, potentially neglecting rare but critical scenarios.

\paragraph{Cross Entropy Importance Sampling (CEIS)} Adopts an optimized proposal distribution $p_\text{CEIS}$ for $ p_\text{training}$, aiming for a balanced approach that covers both common and rare scenarios. The training objective for the ego policy under this approach is formulated as:
\begin{align}
\max_{\pi \in \Pi_\text{ego}} \mathds{E} \left[ \sum_{t=0}^{\infty} \gamma^{t}R(s_t, \pi(s_t)) \cdot \frac{p_\text{naturalistic}(\beta)}{p_\text{CEIS}(\beta)} \right] \label{eq:ego_OEP},
\end{align}
where the importance weight adjusts for the discrepancy between the naturalistic and optimized distributions.

\subsection{Evaluating the Ego Policy}

The evaluation of $\pi_{\text{ego}}$ is designed to mirror realistic conditions, specifically focusing on the policy's effectiveness in dealing with failures such as crashes. We use the CE method iteratively to design the IS proposal distribution $p_{\text{evaluation}}$ optimized for generating challenging scenarios for the corresponding ego policy.

\paragraph{CE Iteration} We initiate the CE algorithm with a Gaussian distribution $N(\mu_0, \sigma^2)$ with some fixed $\sigma$. $\mu_0$ is then iteratively adjusted based on the performance data from the lowest 10th percentile reward of the simulated scenarios, ensuring focus on scenarios that reveal potential weaknesses in the ego policy. In each iteration, a handful of $\beta$ values are sampled from this distribution to simulate driving scenarios that critically evaluate $\pi_{\text{ego}}$. This iterative optimization process is repeated until the parameters of the distribution coverges, indicated by an infinitesimal change in subsequent $\mu$'s. 

\paragraph{Final Metric} To quantify the effectiveness of the ego policy, we compute the final evaluation metric as
\begin{align}
\hat \phi_{\pi_{\text{ego}}} = \frac{1}{N_s} \sum_{i=1}^{N_s} \mathds{1}\{\text{failure}_i \mid \pi_{\text{ego}}\} \frac{p_{\text{naturalistic}(\beta_i)}}{p_{\text{evaluation}(\beta_i)}},\label{eq:eval_IS}
\end{align}
where $N_s$ is the number of samples generated from $p_\text{evaluation}$. This metric provides an unbiased estimate of the naturalistic failure rate:
\begin{align}
\mathds{E}_\beta \left[\mathds{1}\{\text{failure} \mid \pi_{\text{ego}}\}\right] = \mathds{E} \left[ \hat \phi_{\pi_{\text{ego}}} \right].
\end{align}
This approach ensures that although the scenarios are generated from a biased distribution $p_\text{evaluation}$, the final performance estimate remains unbiased.

\subsection{Augmenting Training Distribution}

To refine the training of the ego policy $\pi_{\text{ego}}$, we use a Gaussian Mixture Model (GMM) for the training distribution. The GMM uses parameters derived from the collection of IS proposal distributions generated throughout the evaluation phase. The mean vector of the GMM comprises all the means from the distributions $p_{\text{evaluation}}$, and the standard deviation vector $\sigma$ values. We assign equal weights to each component of the mixture, represented by $1/k$, where $k$ is the number of ego policy training iterations.

This method efficiently uses the diverse and specific scenarios identified during the evaluation phase to enhance the training environment. In addition, the use of the IS-based reward strategy guarantees that the training process yields an unbiased estimate of the ego policy's performance under naturalistic driving conditions and ensures that the modifications made during the training phase lead to performance improvements, \revision{given that the challenging scenarios sampled by the IS proposal can eventually be learned by the ego policy. Otherwise, the training process will not be able to improve the policy and might diminish the performance. To avoid this issue, in the implementation, we filter out the scenarios that are not learnable by the ego policy after a few iterations and reweight the GMM weights accordingly to explore different regions of the scenario space.} The framework is summarized in Algorithm \ref{alg:ego_policy}.

\begin{algorithm}[!tbp]
    \caption{IS-Guided Meta Training}
    \label{alg:ego_policy}
    \begin{algorithmic}[1]
    \State \textbf{Input:} $p_\text{naturalistic}, p_0, \mu_0, \sigma, \Bar B, N_s, K, \beta_\text{min}, \beta_\text{max}$ 
    
    \State \textbf{Meta Policy:}
    \For{each $\Bar{\beta}^i$ in $\Bar{B}$}
        \State $\pi^*_{\text{social},\Bar{\beta}^i}= \text{SocialTraining}(\Bar \beta^i)$ (\autoref{eq:social_objective})
    \EndFor
    
    \Repeat
        \State Sample $\beta \sim U(\beta_\text{min}, \beta_\text{max})$
        \State Regularize $\pi_{\text{social}, \beta}$ to closest $\pi^*_{\text{social}, \Bar{\beta}^i}$ (\autoref{eq:reg_loss})
    \Until{convergence}
    
    \State \textbf{Ego Policy:}
    \For{$k \in 1:K$}
    \State $p_\text{optimized} = p_{k-1}$
    \State $\pi^*_{\text{ego}, k} = \text{EgoTraining}(\pi_{\text{social}, \beta}, p_\text{optimized})$~\autoref{eq:ego_OEP}
    
    \State $\mu^*_k = \text{CE}(\mu_{k-1}, \sigma, \pi^*_{\text{ego}, k})$
    
    \State $p_\text{evaluation} = N(\mu^*_k, \sigma^2)$
    
    \For{$i = 1$ to $N_s$}
        \State Sample $\beta_i \sim p_\text{evaluation}$
    \EndFor
    \State Compute $\hat \phi_{\pi_\text{ego}, k}$  (\autoref{eq:eval_IS})
    
    \State $p_k = \text{GMM}\left([\mu^*_j]_{j \in[k]}, [\sigma, ...], [1/k, ...]\right)$
    \EndFor
    \State \textbf{Output:} Ego policy~$\pi^*_{\text{ego}, K}$
    \end{algorithmic}
    \end{algorithm}

\vspace{-0.1cm}
\section{Experiments}\label{sec:experiments}

\revision{We evaluated our iterative training approach (CEIS) against three baseline methods: generalized ego policy (GEP), naturalistic ego policy (NEP), and generalized with importance sampling (GIS). Our experiments assess how well each method handles both synthetic and real-world driving scenarios.}

\revision{To ensure rigorous evaluation, we implement a careful data partitioning strategy. For the ego vehicle, we maintain separate training and testing datasets. The social vehicles train on the complete dataset, while the ego policy follows a more structured approach. For single-iteration methods, we use the full training set. Our iterative approaches require an additional split of the training data - one portion for policy training and another for intermediate evaluation. This intermediate set helps construct both $p_\text{evaluation}$ and $p_\text{training}$, which guide the policy updates. By maintaining this strict separation between training and testing data, we ensure the final performance metrics for the ego policy remain unbiased.}

\revision{For training the social vehicle, we set the primary reward $r^1$ to correspond to navigating the intersection or roundabout without collision, and the secondary reward $r^2$ to maintain high speed. We use $\overline{B} = \{-1, -0.5, \cdots, 2.5, 3\}$ for the guiding policies, where lower values of $\beta$ corresponds with more aggressive driving behaviors. For anchoring the social policy to the guiding policies, we set the threshold $d = 0.5$. Finally, for statistical robustness, we repeat each experimental configuration 10 times and report both means and standard deviations.}

\subsection{Synthetic Naturalistic Distribution Example}

\revision{For our initial study, we create an intentional mismatch between naturalistic and training distributions.} We set $p_\text{naturalistic} = N(1.5, 0.5^2)$ and $p_\text{training} = N(0.5, 0.5^2)$, deliberately exposing the ego policy to more challenging scenarios during training than it would encounter in typical operation. In this setup, we implement four training schemes: GEP and GIS trained using uniform distributions, NEP trained on the naturalistic distribution, and CEIS trained using our proposed training distribution ($K=1$, non-iterative). 
Each policy undergoes extensive testing with 5,000 simulation runs. During evaluation, we sample $\beta$ from a CE-optimized distribution ($p_\text{evaluation}$) constructed from isolated test data. \revision{We compute final metrics using importance sampling with a naturalistic distribution fitted to the test set and summarize the means and standard deviations of the success rate (SR), collision rate (CR), and timeout rate (TR) in Table \ref{tab:ego_comparison_synthetic}.} 

We also conduct an ablation study, testing Gaussian training distributions with parameters $\mu \in \{-0.5, 0.5, 1.5, 2.5\}$ and $\sigma \in \{0.5, 0.75, 1.0\}$, summarized in Table \ref{tab:ego_ablation_synthetic}, to evaluate whether the CE optimized training distribution leads to maximal performance.

\begin{table}
    \centering
    \caption{Ego policies evaluation for the synthetic dataset }\label{tab:ego_comparison_synthetic}
    \setlength{\tabcolsep}{1.7mm}
    \begin{tabular}{m{0.9cm}<{\centering} | m{1.7cm}<{\centering} | m{1.4cm}<{\centering} m{1.3cm}<{\centering} m{1.3cm}<{\centering}}
        \toprule
        \textbf{Ego Policy} & \textbf{Training Distribution} & \textbf{SR $\uparrow$ ~ \revision{(avg$\pm$std)}} & \textbf{CR $\downarrow$ ~ \revision{(avg$\pm$std)}} &\textbf{TR $\downarrow$ ~ \revision{(avg$\pm$std)}}  \\
        \midrule
        GEP& $U(-1, 3)$ & 0.66$\pm$0.47 & 0.33$\pm$0.47 & \textbf{0.01$\pm$0.08} \\
        GIS& $U(-1, 3)$ & \textbf{0.88$\pm$0.34} &\textbf{0.11$\pm$0.31} & 0.10$\pm$0.13 \\
        NEP& $N(1.5, 0.5^2)$ & 0.68$\pm$0.47 & 0.12$\pm$0.14 & 0.20$\pm$0.32 \\
        CEIS& $N(0.5, 0.5^2)$ & 0.81$\pm$0.39 & 0.11$\pm$0.39 & 0.08$\pm$0.31 \\
        \bottomrule
    \end{tabular}
\end{table}

\begin{table}
    \centering
    \caption{Ego policy results for the synthetic example trained using Gaussian with various means and standard deviations}\label{tab:ego_ablation_synthetic}
    \setlength{\tabcolsep}{1mm}
    \begin{tabular}{m{1.2cm}<{\centering} | m{1.2cm}<{\centering} | m{1.2cm}<{\centering}  c c c}
        \toprule
        \diagbox{\textbf{Std}}{\textbf{Avg}} & \textbf{Metric \revision{(avg$\pm$std)}} & $\mu=-0.5$ & $\mu=0.5$ & $\mu=1.5$ & $\mu=2.5$ \\
        \midrule
        \multirow{3}{*}{$\sigma=0.5$} & SR $\uparrow$  & 0.76$\pm$0.42 & \textbf{0.81$\pm$0.39} & 0.68$\pm$0.47 & 0.70$\pm$0.40 \\
                                      & CR $\downarrow$  &\textbf{0.04$\pm$0.19} & 0.11$\pm$0.39 & 0.12$\pm$0.14 & 0.22$\pm$0.41 \\
                                      & TR $\downarrow$  & 0.20$\pm$0.11 & 0.08$\pm$0.31 & 0.20$\pm$0.32 & 0.03$\pm$0.08 \\
        \midrule
        \multirow{3}{*}{$\sigma=0.75$} & SR $\uparrow$  & 0.77$\pm$0.13 & 0.79$\pm$0.28 & 0.77$\pm$0.31 & 0.78$\pm$0.33 \\
                                       & CR $\downarrow$  & 0.12$\pm$0.15 & 0.17$\pm$0.25 & 0.15$\pm$0.46 & 0.20$\pm$0.31 \\
                                       & TR $\downarrow$  & 0.10$\pm$0.14 & 0.05$\pm$0.13 & 0.08$\pm$0.18 & \textbf{0.02$\pm$0.31} \\
        \midrule
        \multirow{3}{*}{$\sigma=1.0$}  & SR $\uparrow$  & 0.68$\pm$0.49 & 0.70$\pm$0.39 & 0.68$\pm$0.47 & 0.69$\pm$0.31 \\
                                       & CR $\downarrow$  & 0.22$\pm$0.38 & 0.19$\pm$0.39 & 0.21$\pm$0.46 & 0.19$\pm$0.29 \\
                                       & TR $\downarrow$  & 0.10$\pm$0.30 & 0.11$\pm$0.28 & 0.11$\pm$0.27 & 0.12$\pm$0.23 \\
        \bottomrule
    \end{tabular}
    \vspace{-0.4cm}
\end{table}

\subsection{Real-World Naturalistic Distribution}

\revision{In our second experiment, we evaluate our approach against real-world driving behaviors recorded in the InD dataset \cite{inDdataset} and the RoundD dataset \cite{rounDdataset}}. We focus on two T-intersections in Aachen: Heckstrasse and Neukollner Strasse. To extract meaningful behavior parameters, we employ maximum entropy inverse reinforcement learning \cite{huang2021driving} to infer the underlying $\beta$ values from vehicle trajectories. This process assumes our reward structure from Equation (2) and estimates the $\beta$ that best explains each vehicle's observed behavior, with trajectory likelihood proportional to the exponential of the reward.

\begin{figure}
    \centering
    \includegraphics[width=0.9\linewidth]{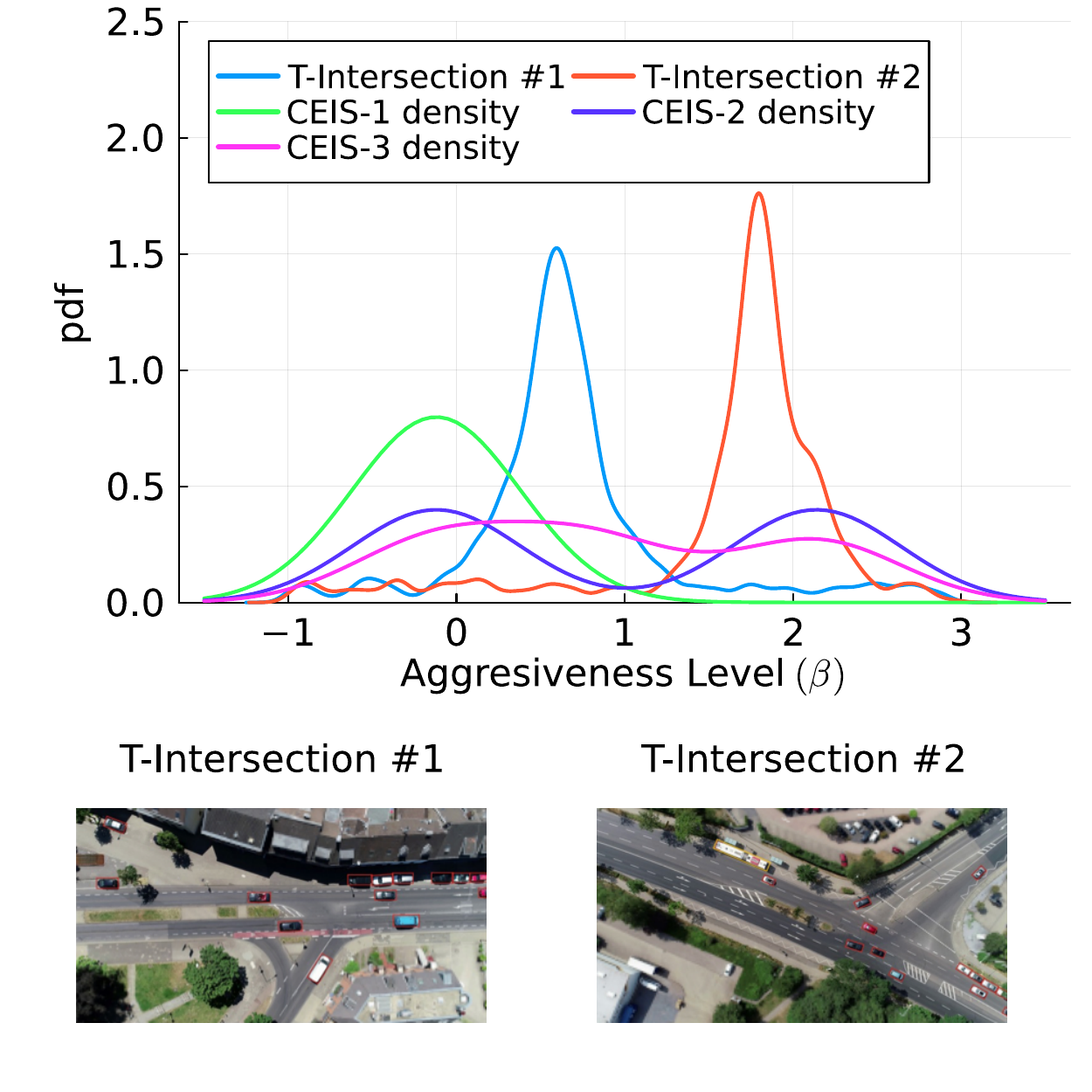}
    \vspace{-0.4cm}
    \caption{Naturalistic and CEIS distributions for the InD T-intersection experiments \cite{inDdataset}.}
    \vspace{-0.4cm}
    \label{fig:naturalistic_pdf}
\end{figure}

Given the estimated $\beta$ samples, we then fit a kernel density estimate to obtain the smoothened density. \autoref{fig:naturalistic_pdf} shows this density for T-intersection \#1 (Heckstrasse, Aachen) and T-intersection \#2 (Neukollner Strasse, Aachen). It is interesting to see that vehicles in T-intersection~\#1 are generally more aggressive (lower $\beta$ values) compared to those in T-intersection~\#2 (higher $\beta$ values), which might be due to the shape of the T-intersection~\#1 that allows turning at faster speed, and thus more quickly crossing the intersection.

Using these naturalistic distributions, we benchmark the same four approaches (GEP, GIS, NEP, and CEIS) as in the synthetic example, with the GEP and GIS using $U(-1, 3)$  and NEP with the kernel density estimate from T-Intersection \#1 as $p_\text{training}$. \revision{We run CEIS for $K=10$ iterations and let it discover an interesting progression of behavior distributions, starting with $N(-0.12, 0.5^2)$, then incorporating $N(2.14, 0.5^2)$ and $N(0.81, 0.5^2)$ in the subsequent two iterations.}

\begin{figure}
    \centering
    \includegraphics[width=0.9\linewidth]{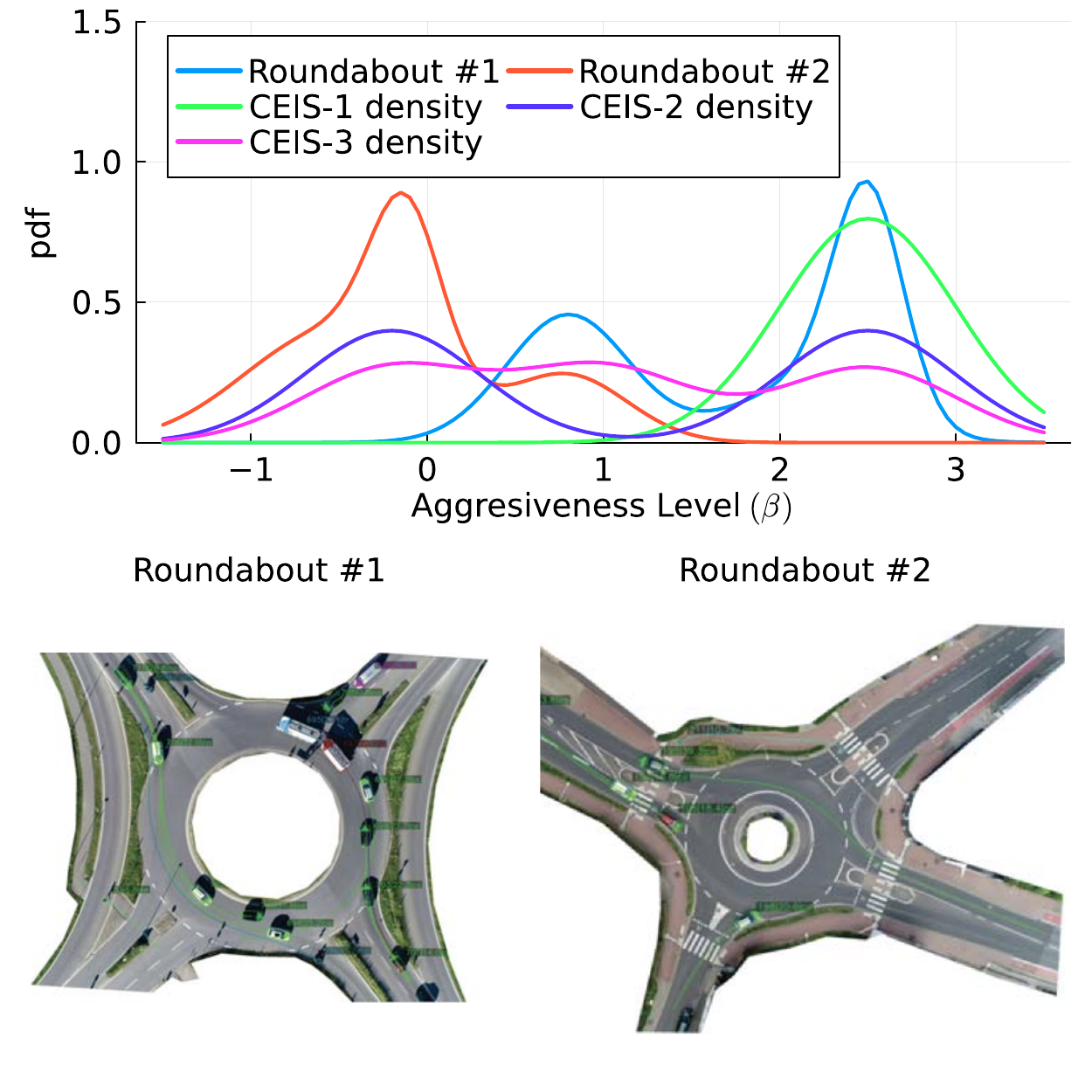}
    \vspace{-0.6cm}
    \caption{\revision{Naturalistic and CEIS distributions for the RoundD roundabout experiments \cite{rounDdataset}.}}
    \vspace{-0.6cm}
    \label{fig:roundabout_pdf}
\end{figure}

\begin{figure}
    \centering
\includegraphics[width=\linewidth]{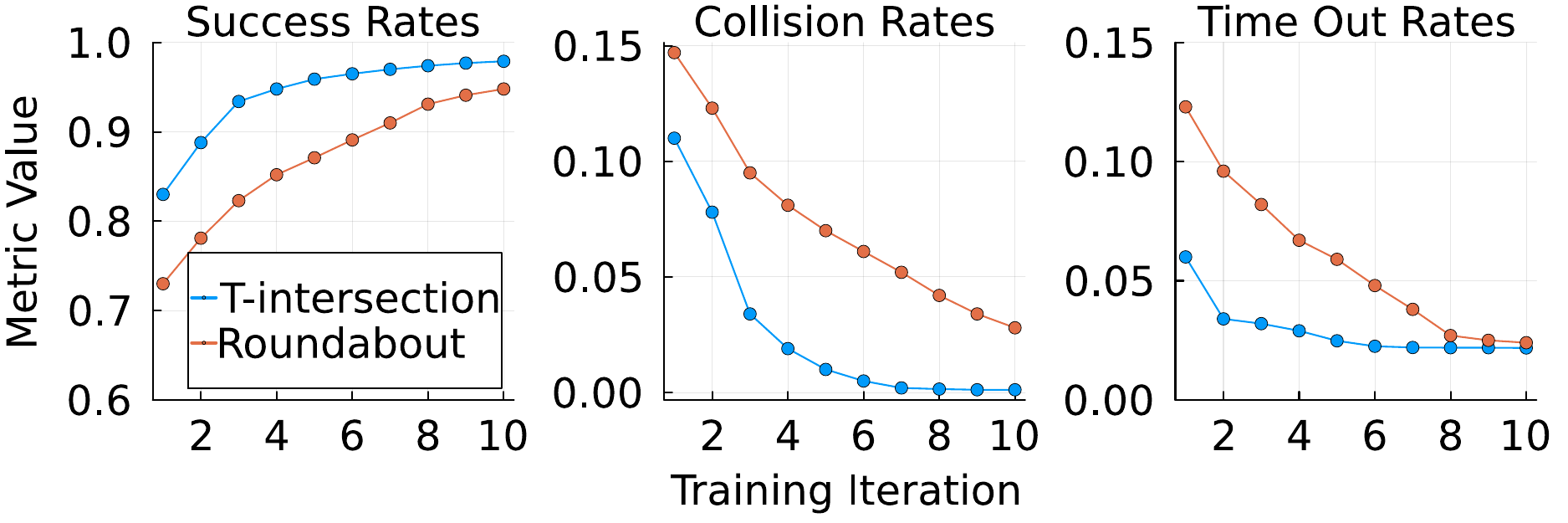}
    \caption{The actual and projected performance for CEIS over longer training iterations.}
    \vspace{-0.5cm}
    \label{fig:oep_metrics}
\end{figure}

\revision{We extend our analysis to include roundabout scenarios from the rounD dataset \cite{rounDdataset}, specifically Neuweilear, Aachen (Roundabout \#1) and Thiergarten, Alsdorf (Roundabout \#2). The training distributions associated with the roundabout dataset are shown in \autoref{fig:roundabout_pdf}, with CEIS using a Gaussian $N(2.55, 0.5^2)$ for iteration 1 and adding $N(-0.24, 0.5^2)$ and $N(0.97, 0.5^2)$ for iterations 2 and 3, respectively. The complete results of our analysis across all scenarios are presented in Table~\ref{tab:ego_comparison_naturalistic} and visualized in \autoref{fig:oep_metrics}. }

\begin{table*}
    \centering
    \caption{Ego policies evaluation for the naturalistic datasets (\revision{InD T-intersections and RoundD roundabouts})}\label{tab:ego_comparison_naturalistic}
    \begin{tabular}{c|m{1.8cm}<{\centering}|m{1.8cm}<{\centering}m{1.8cm}<{\centering}m{1.8cm}<{\centering}|m{1.8cm}<{\centering}m{1.8cm}<{\centering}m{1.8cm}<{\centering}}
        \toprule
        \multirow{3}*{\shortstack[lb]{}} 
		&	& \multicolumn{3}{c|}{\textbf{Intersection \revision{(avg$\pm$std)}}} & \multicolumn{3}{c}{\revision{\textbf{Roundabout (avg$\pm$std)}}} \\
        \cline{3-8}
        \textbf{Ego Policy} & \textbf{Training Distribution} & \textbf{SR} $\uparrow$ & \textbf{CR} $\downarrow$ & \textbf{TR} $\downarrow$ & \revision{\textbf{SR} $\uparrow$} & \revision{\textbf{CR} $\downarrow$} & \revision{\textbf{TR} $\downarrow$} \\
        \midrule
        GEP & $U(-1, 3)$ & 0.75$\pm$0.44 & 0.23$\pm$0.42 & \textbf{0.01$\pm$0.11} & \revision{0.79$\pm$0.31} & \revision{0.15$\pm$0.16} & \revision{0.05$\pm$0.11} \\
        GIS & $U(-1, 3)$ & 0.85$\pm$0.38 & 0.13$\pm$0.34 & 0.02$\pm$0.16 & \revision{0.75$\pm$0.21} & \revision{0.18$\pm$0.10} & \revision{0.07$\pm$0.05} \\
        NEP & KDE & 0.82$\pm$0.39 & 0.08$\pm$0.29 & 0.10$\pm$0.33 & \revision{0.80$\pm$0.22} & \revision{\textbf{0.03$\pm$0.06}} & \revision{0.16$\pm$0.09}\\
        CEIS, $k=1$ & \revision{$N(\mu, \sigma^2)$} & 0.83$\pm$0.38 & 0.11$\pm$0.32 & 0.06$\pm$0.26 & \revision{0.73$\pm$0.15} & \revision{0.15$\pm$0.06} & \revision{0.12$\pm$0.05}\\
        CEIS, $k=2$ & GMM & 0.89$\pm$0.32 & 0.08$\pm$0.29 & 0.03$\pm$0.20 & \revision{0.78$\pm$0.18} & \revision{0.12$\pm$0.07} & \revision{0.10$\pm$0.04}\\
        CEIS, $k=3$ & GMM & \textbf{0.93$\pm$0.28} & \textbf{0.03$\pm$0.19} & 0.03$\pm$0.21 & \revision{\textbf{0.82$\pm$0.16}} & \revision{0.08$\pm$0.04} & \revision{\textbf{0.08$\pm$0.05}}\\
        \bottomrule
    \end{tabular}
    \vspace{-0.3cm}
\end{table*}

\section{Discussion}\label{sec:discussion}


\subsection{Dominant IS-based Performance Over the Baselines}

In both synthetic and real-world experiments, IS-based algorithms demonstrate statistically significant performance advantages over their non-IS counterparts (see \autoref{tab:ego_comparison_synthetic} and \autoref{tab:ego_comparison_naturalistic}). In synthetic scenarios, GIS achieves 88\% success rate and 11\% collision rate, while CEIS achieves 81.1\% success rate and 10.8\% collision rate—representing over 10\% improvement compared to non-IS approaches. The baseline methods show notably weaker performance, with NEP achieving only 68 and GEP 66\% success rates, along with concerning collision rates of 12\% and 33\%, respectively.

\begin{figure}[!tbp]
    \centering
    \begin{subfigure}[t]{0.48\linewidth}
        \centering
        \vspace{-0.65cm}
        \includegraphics[width=\linewidth]{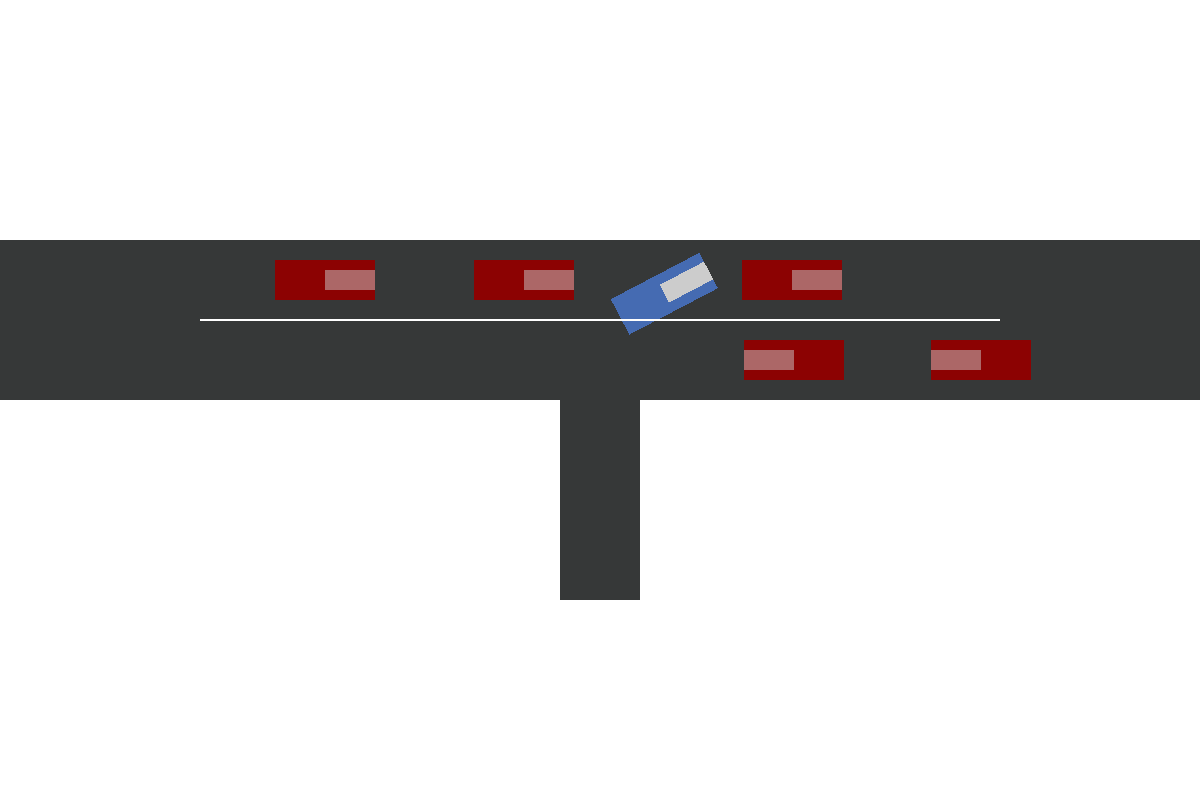}
        \vspace{-1.1cm}
        \caption{Success (the rear vehicle given sufficient reaction time)}
    \end{subfigure}
    ~
    \begin{subfigure}[t]{0.48\linewidth}
        \centering
        \vspace{-0.65cm}
        \includegraphics[width=\linewidth]{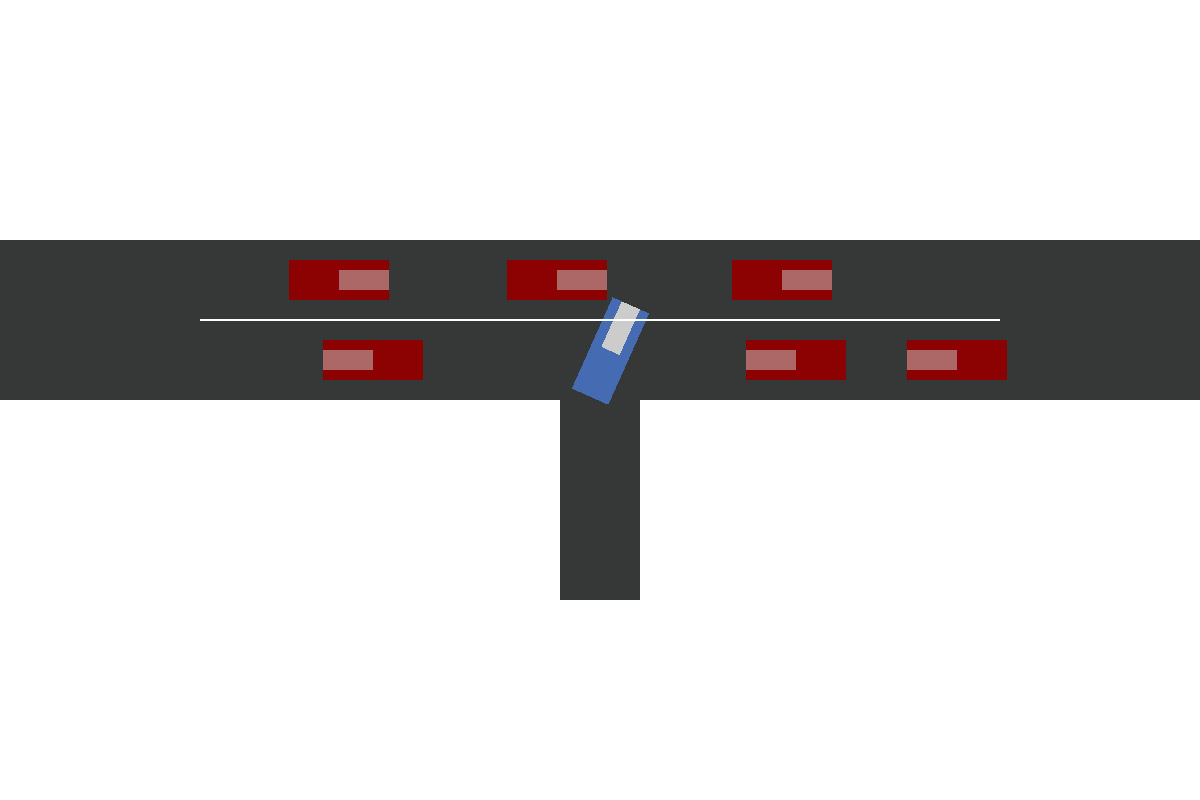}
        \vspace{-1.1cm}
        \caption{Failure (merging initiated too quickly causing a crash)}
    \end{subfigure}
    \vskip\baselineskip
    \begin{subfigure}[t]{0.48\linewidth}        
        \centering
        \vspace{-0.3cm}
        \includegraphics[width=0.9\linewidth]{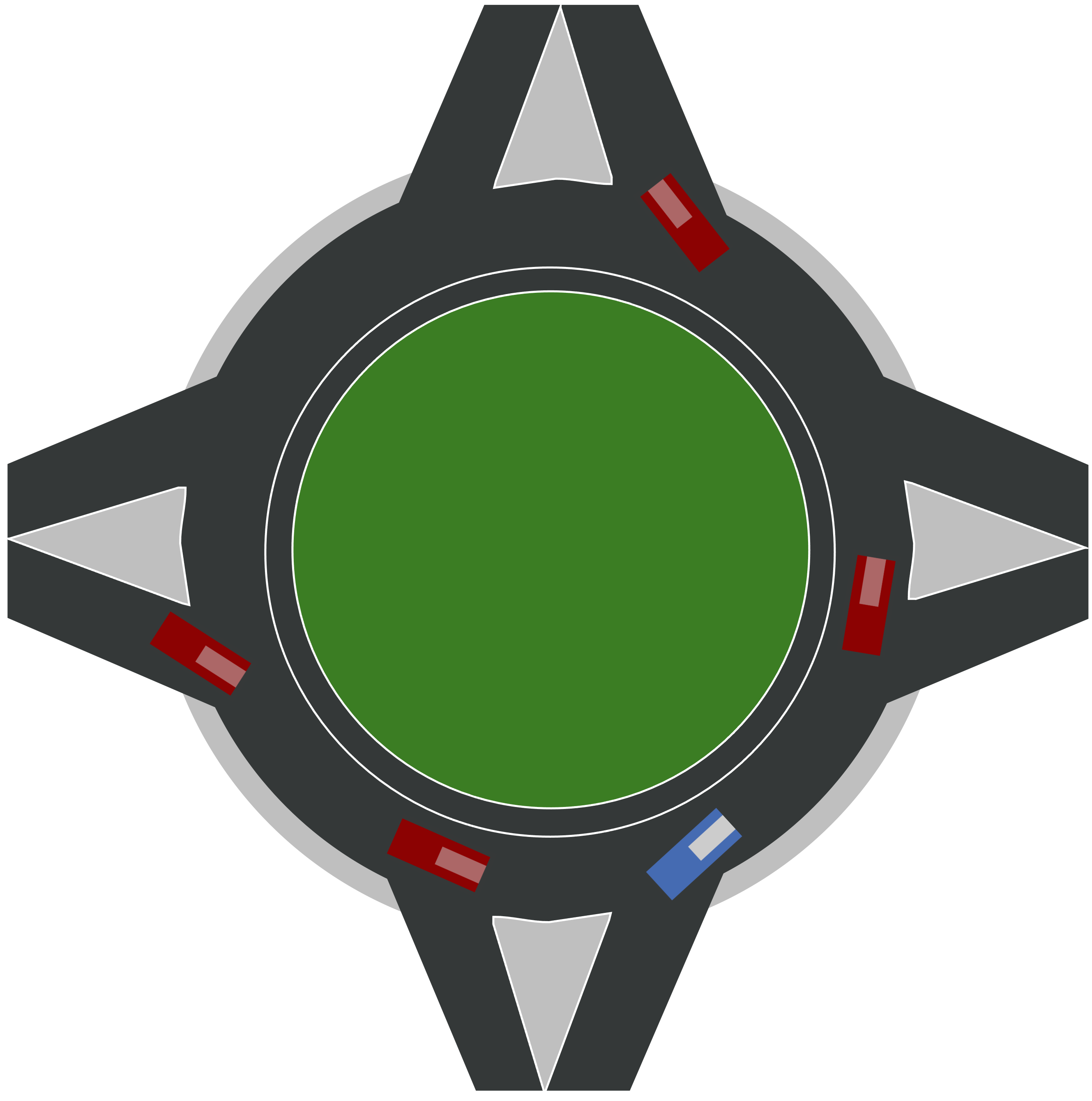}
        \caption{Success (the ego safely navigating the roundabout)}
    \end{subfigure}
    ~
    \begin{subfigure}[t]{0.48\linewidth}
        \centering
        \vspace{-0.3cm}
        \includegraphics[width=0.9\linewidth]{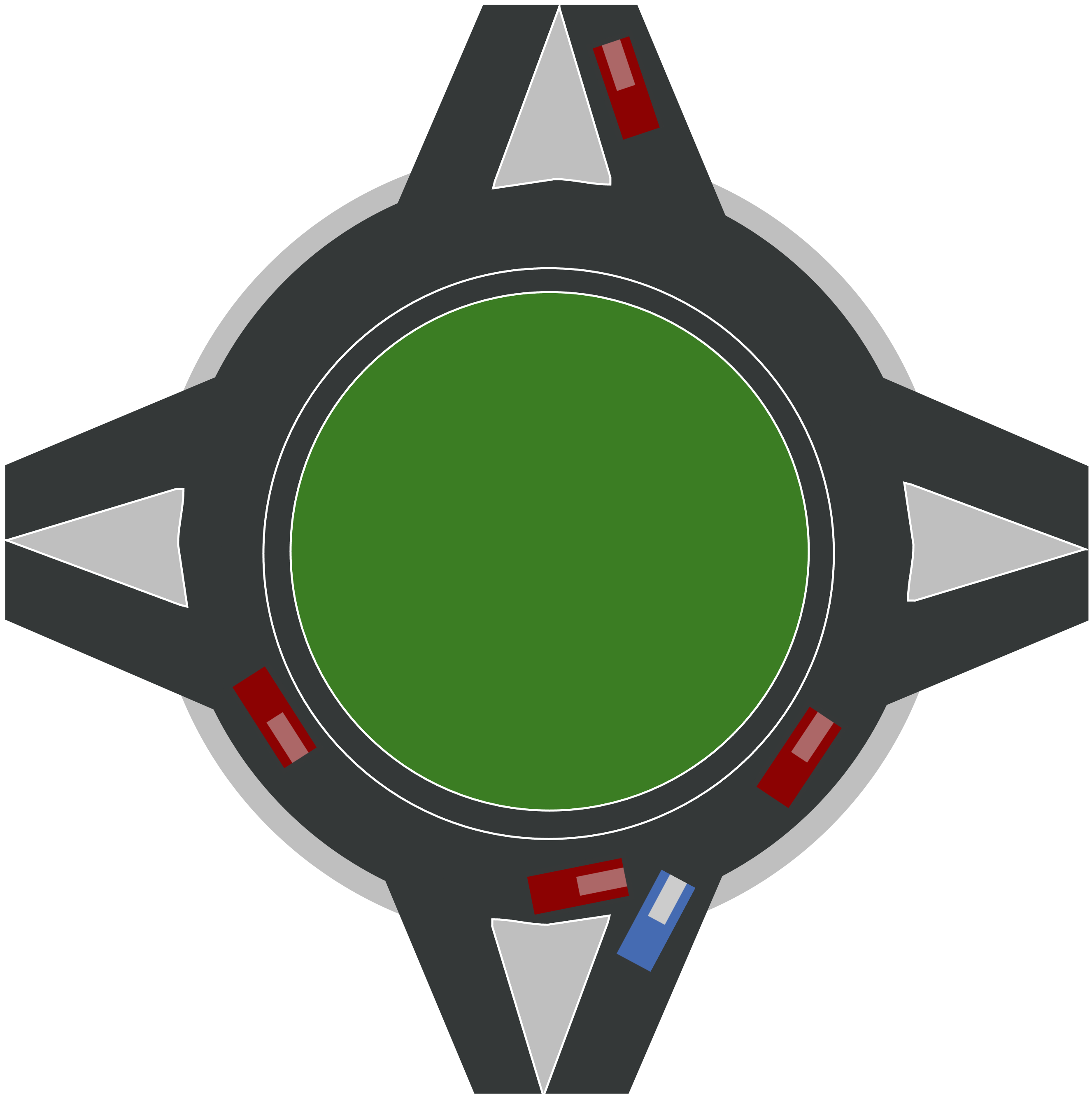}
        \caption{Failure (overly slow merging causing a rear end)}
    \end{subfigure}
    \caption{Success and failure examples of the ego vehicle policy in the T-intersection and roundabout scenarios.}
    \vspace{-0.3cm}
    \label{fig:sim_viz}
\end{figure}

\revision{This performance gap becomes even more pronounced in real-world scenarios, where CEIS achieves a remarkable 93\% success rate in intersections and 82\% in roundabouts by iteration 3 (and 98\% and 94\% by iteration 10, respectively), while maintaining minimal collision rates of 3\% (see \autoref{fig:oep_metrics})}. The only trade-off appears in timeout rates, where GEP shows marginally better performance with 0.6\% timeouts. However, we argue that in practical robotics applications, ensuring safety through careful risk assessment outweighs the marginal benefits of faster task completion. The IS-based approach demonstrates superior risk reasoning by explicitly considering the likelihood of dangerous maneuvers in its decision-making process (as visualized in \autoref{fig:sim_viz}). This represents a key strength of our framework: by systematically evaluating performance across both common and rare scenarios, IS enables more robust adaptation to real-world driving dynamics. The result is an agent that not only excels in typical situations but also maintains reliable performance in unexpected or challenging conditions, as evidenced by the consistently lower standard deviations in CEIS (34\%) compared to GEP (44\%).

\subsection{Better Results in Naturalistic Settings}

While GIS tends to outperform CEIS in the synthetic example, we observe the CEIS' adaptability in the naturalistic case. This is evident from its consistent performance improvements, even when initial training distributions are derived from different intersections (while the GIS slightly underperforms). For instance, when trained with data from diverse intersections, CEIS showed a 6\% increase in success rate over the best baseline, illustrating its ability to generalize well across different traffic scenarios. 

While GIS shows stronger performance in synthetic scenarios (88.0\% success rate), CEIS demonstrates superior adaptability in naturalistic environments. This is particularly evident in its consistent performance improvements across different scenarios types, \revision{achieving a 93\% success rate in intersections and 82\% in roundabouts by iteration 3, while GIS slightly underperforms at 85\% and 75\%, respectively}. The sensitivity analysis in \autoref{tab:ego_ablation_synthetic} reinforces this finding, showing optimal performance with concentrated training distributions ($\sigma=0.5$), especially when combined with CEIS's iterative refinement strategy.

This observation suggests that CEIS's iterative training scheme enables effective continuous learning and adaptation. \revision{However, it is important to note that our current implementation focuses on single-parameter behavioral models ($\beta$), which may not fully capture the complexity of multi-agent interactions in higher-dimensional spaces. However, scaling to more complex scenarios involving multiple types of road users or varying weather conditions would require more sophisticated behavioral models than the current T-intersection and roundabout models used in this study, and thus may require additional research. Nevertheless, these current results demonstrate CEIS' potential as a framework for developing safe and more balanced ego policy to deal with highly interactive environments.}

\subsection{Trend of Improving Performance Over Time}

A longitudinal evaluation of CEIS (\autoref{fig:oep_metrics}) reveals interesting scenario-dependent learning dynamics. In T-intersections, CEIS demonstrates rapid initial improvement—success rates climb sharply from 83\% to 93\% within just three iterations, with an impressive collision rate reduction from 11\% to 3\%. \revision{The learning curve for roundabouts, while positive, shows more modest gains from 73\% to 82\% over the same period, with collision rates decreasing from 15\% to 8\%. This  different learning rates suggest important insights about scenario complexity. T-intersections show approximately 5\% improvement per iteration in the first three iterations, compared to 3.4\% for roundabouts. Moreover, the timeout rates exhibit particularly interesting behavior: they converge much faster in T-intersections (dropping below 3\% by iteration 2) than in roundabouts (requiring 4-5  or even more iterations), suggesting different risk-learning dynamics and that CEIS learns to balance safety and progress more efficiently in simpler geometric configurations.}

\begin{figure}    
    \begin{subfigure}{0.48\linewidth}
        \centering
        \includegraphics[width=\linewidth, trim={1mm 1.5cm 1mm 1.5cm},clip]{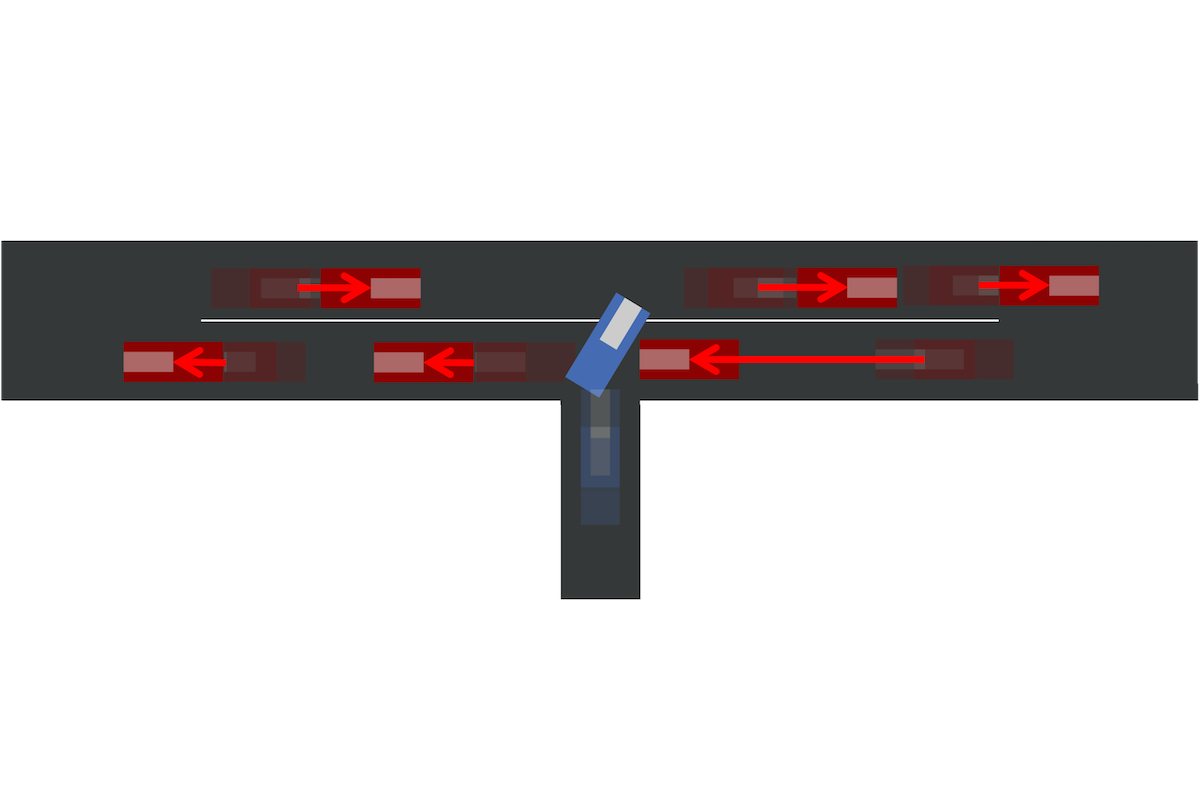}
        \caption{Sudden acceleration.}
    \end{subfigure}
    ~
    \centering
    \begin{subfigure}{0.48\linewidth}
        \centering
        \includegraphics[width=\linewidth, trim={1mm 1.5cm 1mm 1.5cm},clip]{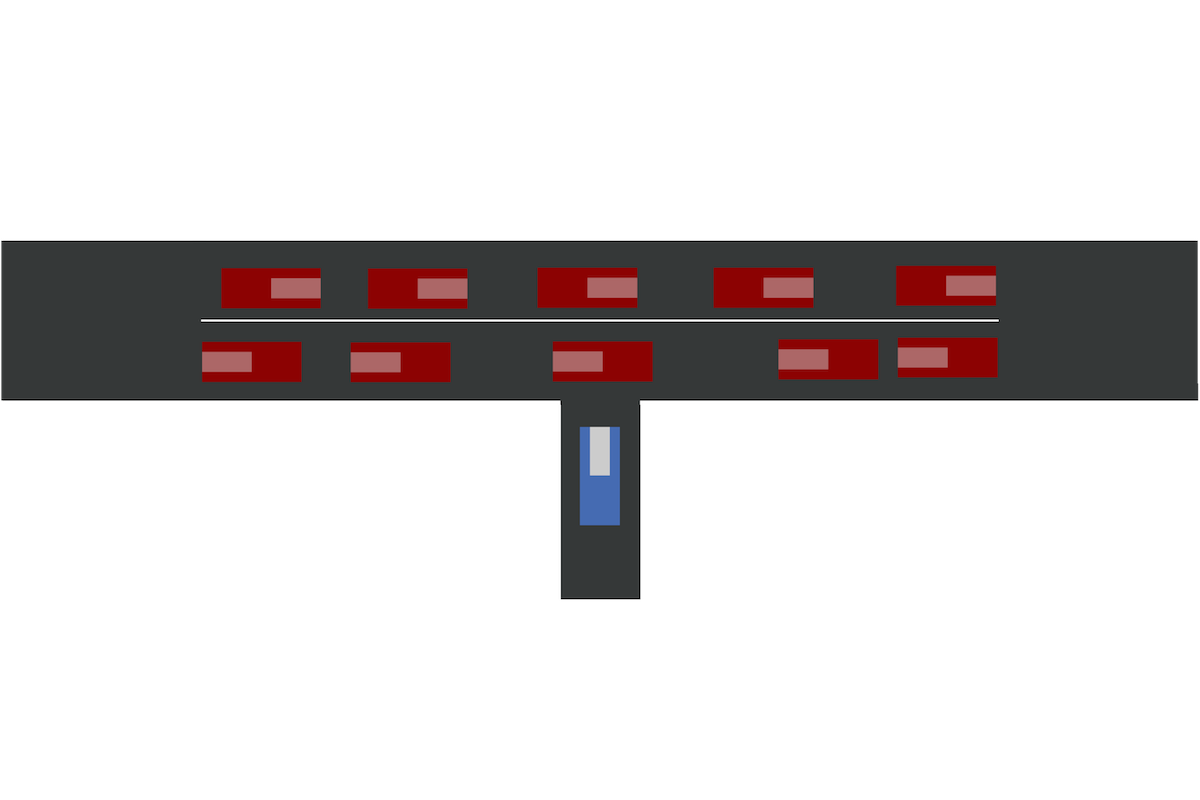}
        \caption{Overcrowded scene.}
    \end{subfigure}
    \caption{Unavoidable failure cases sampled by CEIS proposal.}
    \vspace{-0.3cm}
    \label{fig:unavoidable_failures}
\end{figure}

\revision{More importantly, however, a critical limitation emerges: both scenarios eventually hit performance plateaus (98\% vs 94\% projected success rates), albeit at different iteration counts. This suggests that CEIS's current formulation might struggle with scenario-specific `irreducible complexity',i.e. cases where the inherent multi-agent interaction patterns create unavoidable failure modes (see \autoref{fig:unavoidable_failures} for two such examples from the intersection experiments). This observation opens exciting research directions, particularly in developing scenario-aware training strategies that could dynamically adjust sampling distributions based on detected complexity levels, potentially breaking through these scenario-specific performance ceilings.}

\subsection{\revision{Generalization and Scalability}}

\revision{Our framework demonstrates robust generalization capabilities across diverse driving scenarios, supported by concrete performance metrics. In synthetic environments, CEIS achieves 81\% success rate, while in real-world settings it reaches 98\% for intersections and 94\% for roundabouts. This consistent performance spans not only across scenario types but also across behavioral distributions—from simplified synthetic models to complex real-world patterns extracted from the InD and RounD datasets, suggesting effective transfer of learned safety behaviors across domains.}

\revision{However, scalability presents notable challenges. While CEIS shows computational efficiency in current scenarios, scaling to higher-dimensional problems faces two key limitations: the growing computational cost of CE optimization and the increasing complexity of GMM fitting in larger parameter spaces. Additionally, extending to multi-modal sensing scenarios (cameras, LiDARs, IMUs) would require sophisticated noise modeling and distribution estimation techniques. These scalability challenges point to promising research directions, particularly in developing efficient alternatives to CE optimization \cite{arief2021deep,deo2023achieving} and exploring adaptive GMM architectures for high-dimensional spaces. Future work might also investigate transfer learning approaches to reduce the computational burden when adapting to new sensor modalities or scenario types.}

\section{Conclusion}\label{sec:conclusion}
This research investigates a framework that integrates IS approaches in both the training and evaluation of autonomous agents under complex interactive scenarios. By integrating guided meta reinforcement learning with CE through the IS framework, the training approach adapts efficiently to both common and rare driving conditions. The experimental results using both synthetic and real-world data highlight the potential for substantial improvements in agent performance, with success rates improving significantly across iterative training sessions. Our findings highlight compelling evidence of the potential to integrate IS framework in both training and evaluation to enhance the performance of the ego agents over time. \revision{Future work will focus on exploring the scalability of the proposed framework to more complex multi-agent scenarios and sensor modalities, as well as developing adaptive training strategies to address scenario-specific performance ceilings.}


\printbibliography

\end{document}